\documentclass[11pt]{article}

\usepackage{amsmath,amsthm,verbatim,amssymb,amsfonts,amscd, graphicx}
\usepackage{plstx}
\usepackage{graphics}
\theoremstyle{plain}

\theoremstyle{definition}
\newtheorem{definition}{Definition}
\usepackage{stmaryrd}
\usepackage{alltt}
\usepackage[table]{xcolor}
\usepackage{colortbl}
\usepackage{graphicx}
\usepackage{multirow}

\usepackage{authblk}
\title{This is some thing}
\author[]{Kyle Richardson\thanks{Institute for Natural Language Processing, University of Stuttgart, Germany. Thanks to Richard Waldinger and Jonathan Berant for feedback on an earlier draft.}}
\date{}                     
\usepackage[round,compress]{natbib}
\begin{document}
\title{A Language for Function Signature Representations}
\maketitle

\begin{abstract}

Recent work by \citep{richardson:17a,richardson:17b,richardson:18a}   looks at semantic parser induction and question answering in the domain of source code libraries and APIs. In this brief note, we formalize the representations being learned in these studies and introduce a simple domain specific language and a systematic translation from this language to first-order logic. By recasting the target representations in terms of classical logic, we aim to broaden the applicability of existing code datasets for investigating more complex natural language understanding and reasoning problems in the software domain.

\end{abstract}

\begin{section}{Introduction}

Recent work in natural language processing has looked at learning text to code translation models using parallel pairs of text and code samples from example source code libraries (for a review, see  \cite{neubig2016survey}). In particular, \cite{richardson:17a,richardson:17b,richardson:18a} look at learning to translate short text descriptions to function signature representations as a first step towards modeling the semantics of function documentation. Examples pairs of \emph{docstring} and \emph{function signature} representations are shown in Figure 1; using such pairs, the goal is to learn a general model that can robustly translate a given description of a function to a formal representation of that function. 

Initially, these datasets were proposed as a synthetic resource for studying semantic parser induction \citep{mooney2007learning}, or for building models that learn to translate text to formal meaning representations from parallel data (see \cite{richardson2017code2text} for a proposal on using these datasets for the inverse problem of data-to-text generation). To date, we have built around 45 API datasets across 11 popular programming languages (e.g., Python, Java, C, Scheme, Haskell, PHP) and 7 natural languages (see \cite{RichardsonData}), each using an ad hoc rendering of the target function signature representations.  In this brief note, we define a unified syntax for expressing these representations, as well as a systematic mapping into first-order logic and a small subject domain model.  In doing this, we aim to answer the following question: what do these function signatures that are being learned actually mean, and how can they be used for solving more complex natural language understanding problems (for a similar idea, see \cite{bos2016expressive})?

By recasting the learned representations in terms of classical logic, the hope is that our datasets  will in particular  be made more accessible to studies on natural language based program synthesis \citep{raza2015compositional} and natural language programming  more generally. In what follows, we first define a general syntax for these representations, then discuss the mapping into  logic and the various applications that motivate our particular approach and subject domain model. 

\end{section}

\begin{section}{A Unified Syntax for Function Signatures}
\begin{definition} \textbf{(Syntax of Function Signatures)}\\
$\alpha ::= \texttt{l} \thinspace\thinspace \texttt{N} \thinspace\thinspace \texttt{C}::\texttt{f}(\texttt{t$_{1}$:p$_{1}$},...,\texttt{t$_{n}$:p$_{n}$})  \texttt{ -> r}$
\end{definition}
\noindent As shown in Figure 1, function signature representations across different programming languages  consist of the following components: a \emph{namespace} \texttt{N} (indicating the position or path in the target API), a \emph{class} or \emph{local name identifier} \texttt{C}, a \emph{function name} \texttt{f}, a sequence of (optionally typed \texttt{t}) \emph{named parameters} \texttt{p}, and an (optional) \emph{return value} \texttt{r}. Below shows the different parts of the Java \texttt{max} function: 
\begin{align*}
\underbrace{\texttt{lang}}_{\texttt{N}}  \thinspace\thinspace  \underbrace{\texttt{Math}}_{\texttt{C}}  \thinspace\thinspace \underbrace{\texttt{long}}_{\texttt{r}} \thinspace\thinspace  \underbrace{\texttt{max}}_{\texttt{f}} \texttt{( }  \underbrace{\texttt{long a, long b )}}_{\texttt{t$_{1}$ p$_{1}$,  t$_{2}$, p$_{2}$}}
\end{align*}
In languages or software projects where some of this information is missing, we can mark the positions using special tokens, such as \texttt{UNK}, or unknown, for types in dynamically typed languages, or \texttt{core} and \texttt{builtin} in cases where the namespace and class information are missing. In Definition 1, we define a generic syntax for function signature representations in order to  eliminate superficial differences between different programming languages. This definition includes an additional token $\texttt{l}$ that identifies the particular programming language or software project from which the function \texttt{f} is drawn (see Figure 2 for a \emph{normalized} version of our Java example).

\begin{figure}
\centering 
\footnotesize
\begin{tabular}{| l l |}
\hline
Docstring & \text{Returns the greater of two long values} \\
Signature & (Java) \texttt{lang Math long max(long a,long b)} \\   \hline
Docstring & \text{Compares two values numerically and returns the maximum} \\   
Signature  & (Python) \texttt{decimal Context max(a b)} \\   \hline
Docstring & \text{gibt den gr\"o{\ss}eren dieser Werte zur\"uck} \\   
Signature   & (PHP) \texttt{mixed max(mixed \$value1, mixed \$value2, ..)} \\   \hline
\end{tabular}
\caption{Example function docstring and signature pairs (in a simplified/conventionalized format) for the \texttt{max} function across different programming languages and natural languages. }
\end{figure}

\end{section}
\begin{section}{Semantics and Translation to Logic}

In order to provide a model theoretic semantics of these signature representations, we define a systematic mapping from  $\alpha$ to logic. We also use a small inventory of domain specific predicates to define the semantics, which are motivated by some of the applications that we discuss in the concluding section. 

\paragraph{Function and Return Values} Definition 2 shows the semantics of general function signatures:
\begin{definition}{(\textbf{Function Semantics})} \\
$\llbracket \texttt{l N C::}\texttt{f(t$_{1}$:p$_{1}$,...,t$_{n}$:p$_{n}$) -> r} \rrbracket  = \hspace{10cm}$ 
\\  
$\lambda x_{1}..\lambda x_{n}\exists v\exists f \thinspace\thinspace \texttt{fun}(f,\texttt{f}) \land \texttt{eq}(v,\texttt{f(x$_{1}$...x$_{n}$)})  \land \texttt{lang}(f,\texttt{l}) \land \texttt{type}(v,\texttt{r}) \land \llbracket \texttt{C} \rrbracket \land \llbracket \texttt{N} \rrbracket \land  \llbracket \texttt{t$_{1}$:p$_{1}$} \rrbracket \land ...  \llbracket \texttt{t$_{n}$:p$_{n}$} \rrbracket$
\end{definition}
\noindent The semantics can be described in the following way: for a given function $f$ with some set of function variables $x_{1},..,x_{n}$ (bound here using lambda abstraction), there should exist a value $v$ which is equal  to (shown here using using a special predicate \texttt{eq})  the value that results when the particular function constant \texttt{fun} is applied to said variables.  For example, the variable $v$ in the following example (where lambda conversion is performed on the input 4L, 5L):
\begin{align}
\llbracket \texttt{java lang Math::max(long:a,long:b) -> long} \rrbracket (\texttt{4L})(\texttt{5L})
\end{align}
takes the value of \texttt{5L}, or the result of applying \texttt{max}(4L,5L).  In order to capture additional constraints about typing, naming, and the language from which the function is draw, we use the following domain specific predicates: \texttt{fun} (associates the function variable $f$ with the function constant or name \texttt{f}, e.g., \texttt{max}), \texttt{lang} (the language or project associated with $f$), and \texttt{type} (the type of a given variable, in this case relating the function return variable $v$ with the return type constant \texttt{r}, e.g., \texttt{long}). 


\begin{figure}
\centering 
\footnotesize
\begin{tabular}{| l l |}
\multicolumn{2}{c}{Returns the greater of two long values} \\
\multicolumn{2}{c}{$\Big\downarrow$} \\[.2cm]
\hline
Signature (informal) & \texttt{lang Math long max(long a,long b)} \\ \hline
Normalized & \texttt{java lang Math::max(long:a,long:b) -> long} \\ \hline
& \\[-.3cm]
& $\llbracket$\texttt{java lang Math::max(long:a,long:b) -> long}$\rrbracket$ \\
& \multicolumn{1}{c|}{$\Updownarrow$} \\
Expansion to Logic & 
	\begin{tabular}{l}
	$\lambda x_{1} \lambda x_{2} \exists v \exists f \exists n \exists c  \hspace{.1cm} \texttt{eq}(v,\texttt{max}(x_{1},x_{2})) \land  \texttt{fun}(f,\texttt{max}) $ $\land$ \texttt{type}($v$,\texttt{long})  \\
	\hspace{2cm} $\land$ \texttt{lang}($f$,\texttt{java}) \\
	\hspace{2cm} $\land$  \texttt{var}($x_{1}$,\texttt{a}) $\land$  \texttt{param}($x_{1}$,$f$,\texttt{1})  $\land$ \texttt{type}($x_{1}$,\texttt{long}) \\
	\hspace{2cm} $\land$  \texttt{var}($x_{2}$,\texttt{b}) $\land$  \texttt{param}($x_{2}$,$f$,\texttt{2}) $\land$  \texttt{type}($x_{2}$,\texttt{long}) \\
	\hspace{2cm} $\land$ \texttt{namespace}(n,\texttt{lang}) $\land$ \texttt{in\_namespace}($f$,$n$)  \\
	\hspace{2cm} $\land$ \texttt{class}(c,\texttt{Math}) $\land$ \texttt{in\_class}($f$,$c$)  \\
	\end{tabular}
\\ \hline
\end{tabular}
\caption{An normalized version of the Java example and its translation to logic.}
\end{figure}

\paragraph{Arguments}  Definition 3 shows the semantics of function arguments. 

\begin{definition}{(\textbf{Argument Semantics})} \\
$\llbracket \texttt{t$_{j}$:p$_{j}$} \rrbracket =  \texttt{var}(x_{j},\texttt{p}_{j})  \land \texttt{type}(x_{j},\texttt{t}_{j}) \land  \texttt{has\_param}(f,x_{j},\texttt{j}) $
\end{definition}
\noindent The same naming and typing constraints are expressed using similar predicates for variables. The predicate \texttt{var} associates a given variable assignment $x_{j}$ with an argument name \texttt{p}$_{j}$.  In addition, the predicate \texttt{has\_param} explicitly associates a given argument or parameter and its position with a function $f$. 

\paragraph{Namespace and Classes} Definition 4 shows the semantics of namespaces and classes: 

\begin{definition}{(\textbf{Namespace and Class Semantics})} \\
$\llbracket \texttt{N} \rrbracket = \exists n.  \texttt{namespace}(n,\texttt{N}) \land \texttt{in\_namespace($f$,n)}$ \\
$ \llbracket \texttt{C} \rrbracket = \exists c.  \texttt{class}(c,\texttt{C}) \land \texttt{in\_class}(f,\texttt{c})$
\end{definition}
\noindent Here, we use the predicates \texttt{namespace} and \texttt{class} to identify the type of the variables $n$ and $c$. As with arguments, two additional predicates, \texttt{in\_namespace} and \texttt{in\_class},  are introduced in order to associate particular namespaces and classes  with  particular function values.

Figure 2 shows a full translation from an ad hoc signature representation to a normalized representation and finally to a representation in logic using the definitions introduced above. We note that while we use a specific, and seemingly arbitrary,  set of domain predicates, new predicates and information can be added as needed. In the next section, we motivate the particular predicates chosen above by describing some possible applications of our formulation. 

\end{section}

\begin{section}{Applications and Discussion}

In any application of logic, logical formulas can be used either to reason \emph{extensionally} (i.e., about the particular real-world entities denoted by or involved in a given formula) or \emph{intentionally} (i.e., about abstract relationships and consequences between concepts). Taking the example in Figure 2 and its expansion to logic, we could reason extensionally using pure logic about the exact value that this function will return given a particular input. In contrast, we could also, with the help of additional domain specific knowledge, reason intensionally about abstract relationships between different programming languages, class and namespace structures, and so on. 

%
While we think that there is value in the first type of reasoning, especially for building executable models of functions, our primary focus is on reasoning abstractly about programming language constructs and relationships across different programming languages and projects. One benefit of the source code domain is that much of the declarative knowledge needed for  reasoning can be extracted straightforwardly from the target libraries directly, including information about class containment and subsumption relations, lists of related utilities (e.g., via \emph{see-also} annotations and documentation hyperlinking), function naming alternatives or aliases, and the relative position or distance between different functions and namespaces. Having such knowledge and an expressive logical language can in general facilitate more complex forms of API question-answering and code retrieval (see \cite{richardson:17b}). As an example, we might might use the following notation (in which each $\texttt{v}?$ expands to an existential variable in Definition 2): 
\begin{align}
\texttt{java N? C?::f?(long:a,long:p?) -> long}
\end{align}
to request the following: \emph{Find some java function somewhere (i.e., in some class and namespace), that takes two long values  as arguments (with the first value having the name \texttt{a}) and returns a long value.} Such a request might be used for finding structurally related functions or for mining \emph{software clones} \citep{rattan2013software}.

\begin{figure}

\centering
\scriptsize
\begin{tabular}{| c l l |}
\hline
1. & Source API: (\textbf{\emph{en},\textbf{ Haskell})} & \textbf{Input}: Shift the argument left by the specified number of bits. \\ \hline
\multirow{3}{*}{\rotatebox[origin=c]{90}{Output}} & \multicolumn{1}{l}{Language: \textbf{Haskell}} &  Translation: \texttt{Haskell Data.Bits builtin::shiftL(UNK:a,Int:UNK) -> UNK} \\
& \multicolumn{1}{l}{Language: \textbf{Java}} & Translation: \texttt{Java java.math BigInteger::shiftLeft(int:n) -> BigInteger} \\
&\multicolumn{1}{l}{Language: \textbf{Clojure}} & Translation: \texttt{Clojure clojure.core builtin::bit-shift-left(UNK:x,UNK:n) -> UNK} \\ \hline 
\end{tabular}

\caption{Multi programming language translation output (in a normalized form) for the input \emph{Shift the argument left by the specified number of bits} using the model of \cite{richardson:18a}.}
\end{figure}

Our primary focus is on building models that can robustly translate high-level natural language descriptions to code, and hence to the logical representations  proposed above. We believe that under this scenario, natural language can prove to be a useful tool for  deriving new forms of declarative knowledge. For example, our recent work looks at \emph{polyglot} translation \citep{richardson:18a}, or building text-to-code translators that can translate descriptions to function representations in multiple APIs. An example is provided in Figure 3, where the model translates the description about bit-shifting operations (originally drawn from the \texttt{Haskell} standard library) to equivalent function translations in the \texttt{Haskell},  \texttt{Java} and \texttt{Clojure} standard libraries. With  this output, one could straightforwardly extract  rules about function equivalences in different languages (e.g., \texttt{bit-shift-left} in \texttt{Clojure} is the same function as \texttt{shiftLeft} in \texttt{Java}), and learn further relationships between the associated function names and variables. 

Using the notation introduced above, we can express cross language queries about equivalent functions in the following way: 
\begin{align}
\texttt{java java.math BigInteger::EquivIn(shiftLeft,haskell)(long:a,long:b) -> long}
\end{align}
where the special predicate \texttt{EquivIn} is used to request the \texttt{Haskell} equivalent of the \texttt{shiftLeft} function in \texttt{Java}. The semantics of \texttt{EquivIn} can therefore be defined in the following way (where background knowledge about the \texttt{eq} predicate can be derived from the output of our polyglot model as discussed above):
\begin{align}
\begin{tabular}{c}
$\llbracket$\texttt{l N C::EquivIn(f,lang)(t$_{1}:$p$_{1}$,...,t$_{n}$:p$_{n}$) -> r}$\rrbracket$ \\
$\Updownarrow$ \\
$\llbracket \texttt{l N C::f(t$_{1}:$p$_{1}$,...,t$_{n}$:p$_{n}$) -> r} \rrbracket \land \llbracket$ \texttt{lang N? C?::f'?(?) -> r?} $\rrbracket \land \texttt{eq}(\texttt{f,f'})$
\end{tabular}
\end{align}

One interesting direction is using general knowledge about software libraries and logic reasoning  to help learn more robust translation models. The formalism introduced above is part of an effort to move in this direction, and  we hope that integrating symbolic reasoning more generally will open the doors to new ideas and approaches to solving everyday software search and reusability issues.

\end{section}

\bibliographystyle{plainnat} 
\bibliography{example} 

 \end{document}